\def\ARXIV{}
\documentclass[letterpaper, 10pt, conference]{ieeeconf}

\IEEEoverridecommandlockouts
\usepackage{times}
\usepackage[bookmarks=true]{hyperref}
\hypersetup{hidelinks}
\usepackage{graphicx}
\usepackage{subcaption}
\usepackage{algorithm}
\usepackage{algorithmic}
\usepackage{pifont}
\usepackage{tikz}
\usepackage{multirow}
\usepackage{soul}
\usetikzlibrary{arrows.meta,positioning}

\usepackage{comment}

\newif\ifanon
\anontrue
\ifdefined\ARXIV\anonfalse\fi
\ifanon\else
\hypersetup{pdftitle={EasyNav: A Modular Architecture for Real-Time Navigation Across
  Heterogeneous Robot Environments},
  pdfauthor={Francisco Mart\'in Rico, Francisco Miguel Moreno, Juan Sebasti\'an Cely
  Guti\'errez, Esther Aguado, Miguel \'Angel De Miguel Para\'iso, Jos\'e Miguel Guerrero}}
\fi

\begin{document}

% paper title
\title{EasyNav: A Modular Architecture for Real-Time Navigation Across Heterogeneous Robot Environments}

% Anonymized for double-anonymous review (ICRA 2027). Restore real author block for camera-ready --
% see bitacora.md for the version with names, affiliation, and emails.
\ifanon
\author{Author Names Omitted for Anonymous Review. Paper-ID [add your ID here]}
\else
\author{Francisco Mart\'in Rico, Francisco Miguel Moreno, Juan Sebasti\'an Cely Guti\'errez, \\
Esther Aguado, Miguel \'Angel De Miguel Para\'iso, and Jos\'e Miguel Guerrero% <-this % stops a space
\thanks{All authors are with the Intelligent Robotics Lab, Universidad Rey Juan Carlos, Fuenlabrada,
Madrid, Spain
        {\tt\small \{francisco.rico, franciscom.moreno, juan.cely, esther.aguado,
        miguelangel.demiguel, josemiguel.guerrero\}@urjc.es}}%
}
\fi

\maketitle
\thispagestyle{empty}
\pagestyle{empty}

\begin{abstract}
Most robot navigation stacks implicitly bind mapping, localization, planning, and control to
specific representations and data flows, yielding architectures hard to extend, compare
experimentally, or reason about end-to-end timing. This is critical in heterogeneous scenarios,
where multiple spatial representations, sensing modalities, and coordinate frames coexist and
evolve at different rates.
This paper presents a navigation architecture centered on an explicit, shared navigation state that
serves as the sole interaction interface between modular components. Constraining mapping,
localization, planning, and control to operate over this shared state enables interchangeable,
representation-agnostic pipelines under explicit compatibility and timing contracts. We realize
this architecture as EasyNav, an open source ROS~2 framework, and
validate it through real-robot experiments, including indoor navigation with laser sensing and
large-scale indoor-outdoor navigation with GPS and 3D LiDAR across multi-level routes. Compared
with Nav2 under matched conditions, EasyNav reduces reaction latency from 377 to 33~ms indoors and
from 71 to 8~ms outdoors, while also using less CPU and memory.

\end{abstract}

\section{Introduction}
Autonomous robot navigation has matured into a core capability of modern robotic systems, yet the
architectural principles underlying most navigation stacks have evolved far more slowly than the
sensing, mapping, and learning techniques they integrate. Contemporary
frameworks~\cite{macenski2020marathon}, built on ROS and its successor
ROS~2~\cite{quigley2009ros,macenski_robot_2022}, typically intertwine environment representations,
localization, planning, and control through implicit data flows, ad hoc interfaces, and callback
execution order. This coupling hampers systematic comparison, complicates reuse across robots and
domains, particularly when navigation spans indoor and outdoor environments or combines metric,
semantic, and topological information, and, most critically, leaves end-to-end latency and
time/frame consistency as emergent properties of scheduling rather than design objectives, making
reactivity and determinism hard to guarantee.

\begin{figure}[b]
  \centering

  \begin{subfigure}[t]{0.32\linewidth}
    \centering
    \includegraphics[width=\linewidth]{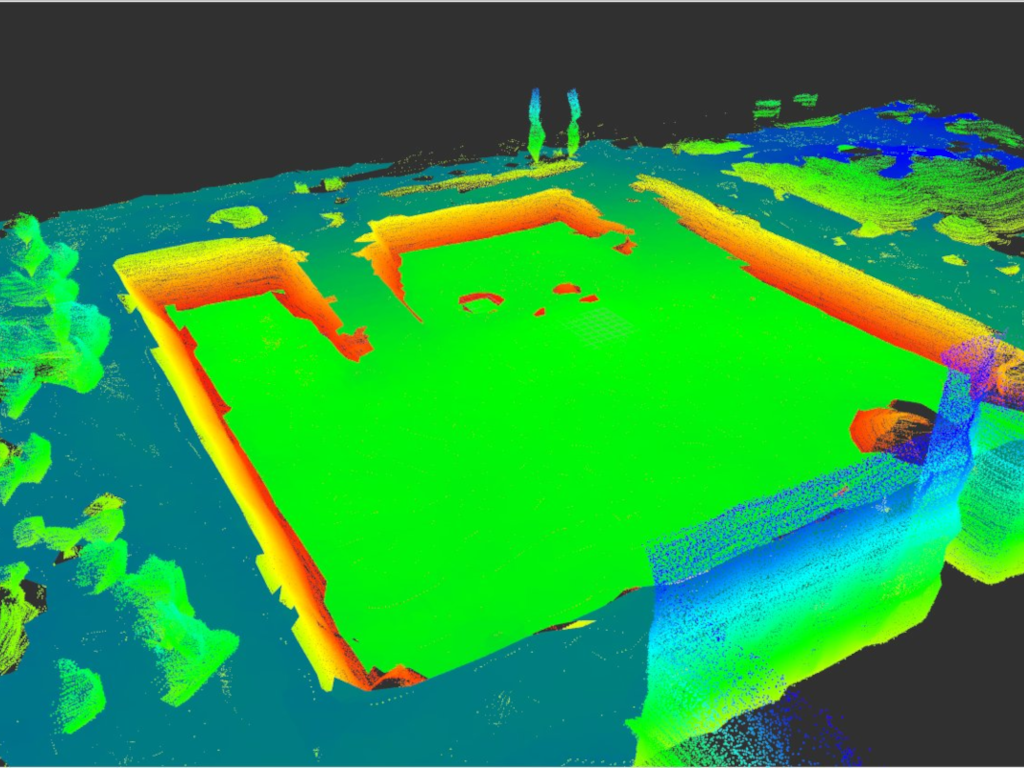}
    \caption{Non-planar outdoor maps composed of navigation surfaces and point clouds.}
    \label{fig:intro:fig1}
  \end{subfigure}\hfill
  \begin{subfigure}[t]{0.32\linewidth}
    \centering
    \includegraphics[width=\linewidth]{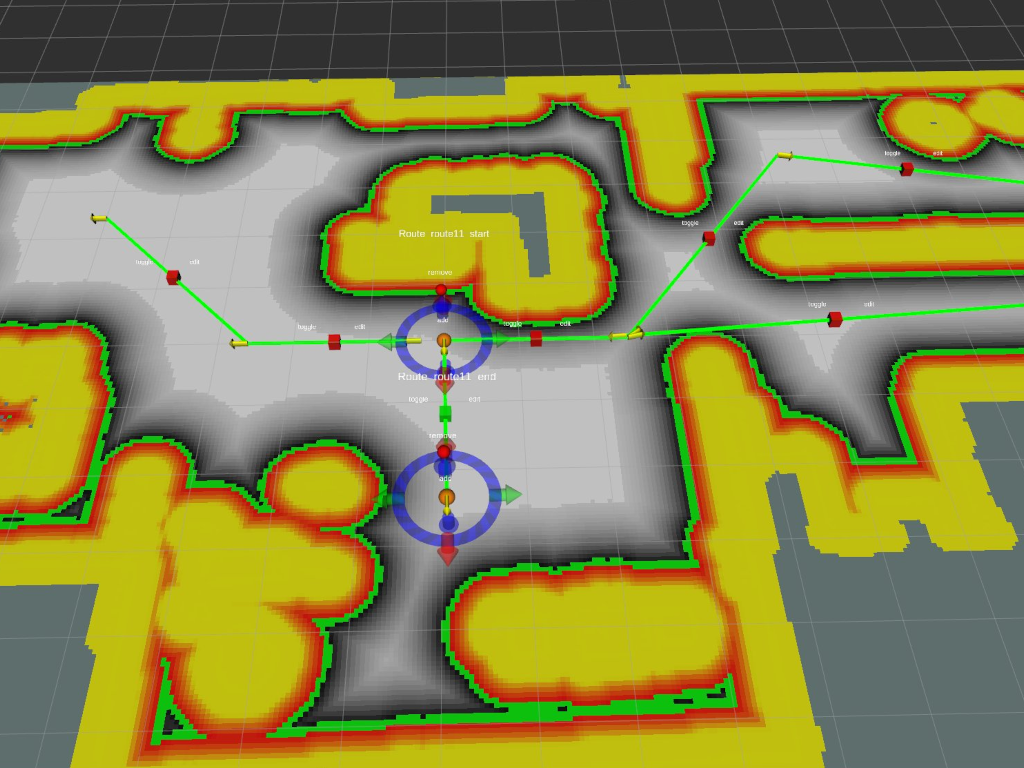}
    \caption{Indoor costmap with inflated obstacles and fixed navigation routes.}
    \label{fig:intro:fig3}
  \end{subfigure}\hfill
  \begin{subfigure}[t]{0.32\linewidth}
    \centering
    \includegraphics[width=\linewidth]{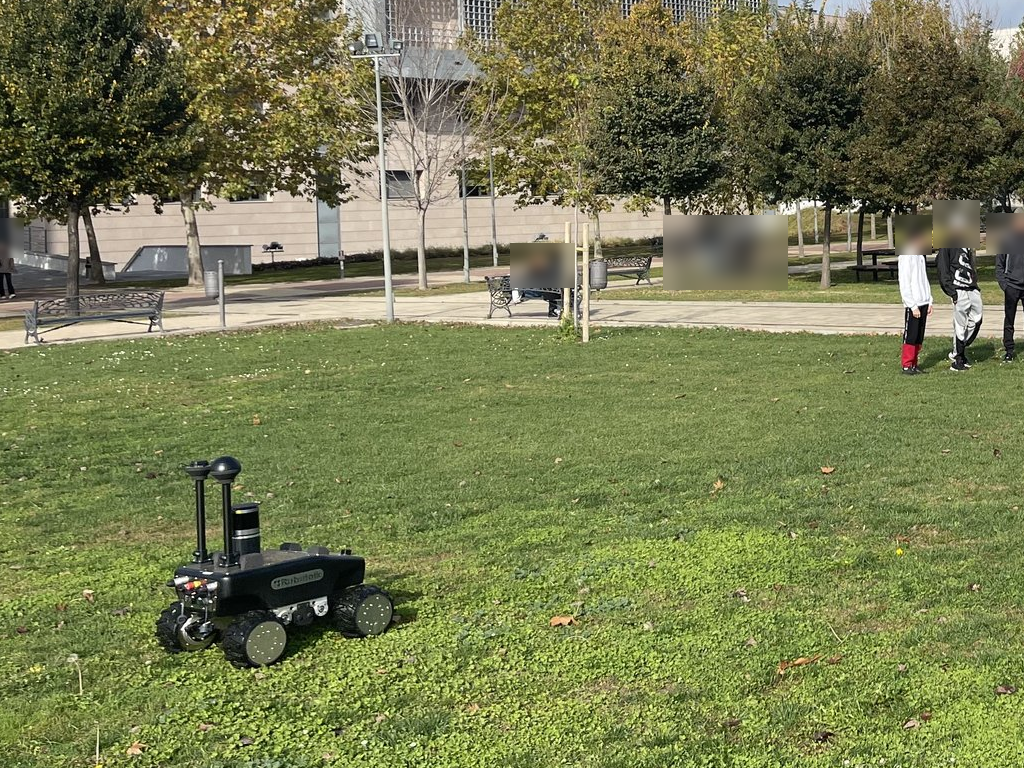}
    \caption{One of the robots used in the experimental evaluation of the proposed architecture.}
    \label{fig:intro:fig4}
  \end{subfigure}

  \caption{Examples of navigation scenarios enabled by the proposed architecture.}
  \label{fig:intro1}
\end{figure}

This paper presents a navigation architecture organized around an explicit navigation state that
captures, in a unified and temporally consistent form, all information required by mapping,
localization, planning, and control. By enforcing interaction exclusively through this shared state
rather than through direct dependencies, navigation components remain modular and interchangeable
under well-defined compatibility and timing constraints, elevating time and frame management to
first-class design concerns and enabling explicitly ordered reactions to new information. We
realize this architecture as EasyNav\ifanon\else\footnote{\url{https://github.com/EasyNavigation}}\fi{}
and validate it on real robotic platforms spanning indoor
navigation with laser sensing and large-scale indoor-outdoor navigation with GPS and 3D LiDAR
across multi-level routes,
showing that architectural clarity and temporal rigor can coexist with practical, real-time
deployment.

This work makes four primary contributions to the design and implementation of robot navigation
systems:
\begin{itemize}
    \item \textbf{An explicit, shared navigation state} through which mapping, localization,
    planning, and control interact exclusively, eliminating implicit data dependencies and
    decoupling navigation functionality from any specific environment representation.
    \item \textbf{Real-time behavior, freshness, and frame consistency as enforced properties} of
    the navigation state, rather than emergent effects of execution order, allowing end-to-end
    latency to be explicitly instrumented, characterized, and controlled at the architectural
    level.
    \item \textbf{Native support for heterogeneous, coexisting environment representations}, such as
    costmaps, multi-level navigable surfaces, and volumetric models, within a single pipeline, with
    components remaining interchangeable as long as they satisfy the shared state's temporal and
    compatibility constraints.
    \item \textbf{EasyNav}, a fully open source, documented, and extensible ROS~2 realization of
    these principles, with a diverse set of ready to use plugins and a low barrier of entry for
    third party contributions.
\end{itemize}

\section{Related Work}
\label{sec:related}

Long before ROS, layered robot control architectures such as LAAS's three-tier
decision/execution/functional design~\cite{alami1998architecture} and JPL's
CLARAty~\cite{volpe2001claraty} already pursued modularity, reusability, and interoperability
across heterogeneous robots and missions through explicit layer boundaries and shared functional
interfaces. CARMEN~\cite{montemerlo2003carmen} took the opposite stance, deliberately avoiding a
strict architectural standard in favor of common conventions to lower the barrier of entry for new
algorithms. EasyNav shares both goals, grounding them in a single, formally declared state contract
rather than layer boundaries or convention alone.

The idea of coordinating independent modules through a shared data store dates back to classical
blackboard architectures~\cite{hayes-roth1985blackboard}, where independent knowledge sources read
and write incrementally to a common workspace without direct knowledge of each other. Robotics
adopted this pattern early: Fawkes~\cite{niemueller2010fawkes} organizes plugins around a
time-stamped blackboard with well-defined read and write interfaces, used as a convenience
mechanism with no declared, checkable manifest of a plugin's full data dependencies. More
recently, behavior-tree engines such as BT.CPP, used by Nav2~\cite{macenski2020marathon}, expose
typed, per-node ports that explicitly declare local data
dependencies~\cite{colledanchise2017behavior}; these ports, however, do not constitute a
navigation-wide contract over validity time, coordinate frame, freshness, or execution ordering.
EasyNav instead makes the shared state the sole interaction interface and attaches
freshness and frame-consistency obligations to every entry, so compatibility between two plugins
becomes an explicit, checkable property rather than an assumption.

Composing several environment representations within one pipeline is not new either: Nav2's layered
costmaps~\cite{lu2014layered} already let independent plugins, such as a static map, obstacle
inflation, or custom layers, contribute sequentially to a single two-dimensional costmap consumed by
the planner. EasyNav generalizes this along two axes: representations are not limited to a single
costmap type, since planar costmaps, multi-level surfaces, and volumetric point clouds may coexist
and feed different consumers, and timestamp/frame propagation is enforced by the architecture
rather than left to each layer's implementation; a similar single-representation commitment appears
in single-scene motion planning~\cite{coleman2014reducing} and single-map terrain
navigation~\cite{fankhauser2018probabilistic}, none designed for heterogeneous representations
under a shared timing discipline.

On the execution side, the Orocos Real-Time Toolkit~\cite{bruyninckx2001orocos} pioneered explicit
separation of real-time and non-real-time components communicating through ports, a goal EasyNav
shares but applies specifically to navigation, coupling it with the time-stamped state above. A
growing body of work targets the default ROS~2 executor's lack of explicit ordering and resulting
priority inversions and starvation, proposing chain-aware schedulers~\cite{choi2021picas} and
refined response-time analyses~\cite{casini2019response}. These bound or reorder the latency of a
fixed callback graph after the fact, at the executor level; EasyNav instead integrates explicit
temporal and frame semantics into the shared navigation state itself, coupling scheduling to the
specific roles, representations, and transformation dependencies of a pipeline rather than a
generic callback chain.
Table~\ref{tab:related} summarizes this comparison.

\begin{table}[t]
\centering
\scriptsize
\renewcommand{\arraystretch}{0.85}
\setlength{\tabcolsep}{3pt}
\caption{Comparison with representative related systems. \ding{51}: supported;
\ding{55}: not supported; \emph{partial}: supported in a restricted sense (see
Section~\ref{sec:related}); n/a: not applicable to that system's scope.}
\label{tab:related}
\begin{tabular}{p{1.9cm}ccccc}
\hline
\textbf{Property} & \textbf{Fawkes} & \textbf{Nav2} & \textbf{Orocos} & \textbf{CLARAty} &
\textbf{EasyNav} \\
& \cite{niemueller2010fawkes} & \cite{macenski2020marathon,colledanchise2017behavior} &
\cite{bruyninckx2001orocos} & \cite{volpe2001claraty} & \\
\hline
Global nav-state R/W contract & partial & \ding{55} & n/a & n/a & \ding{51} \\
Per-entry stamp/frame & partial & \ding{55} & n/a & n/a & \ding{51} \\
Heterog. repr., 1 pipeline & n/a & partial & n/a & n/a & \ding{51} \\
Perception-triggered exec. & \ding{55} & \ding{55} & partial & n/a & \ding{51} \\
Explicit RT/non-RT split & \ding{55} & \ding{55} & \ding{51} & partial & \ding{51} \\
\hline
\end{tabular}
\end{table}

\section{Architecture Overview}
\label{sec:arch}
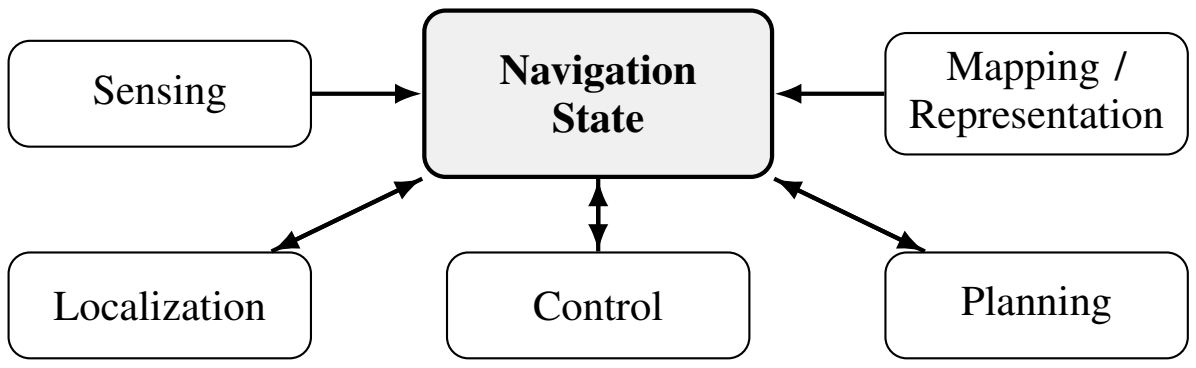
\begin{figure}[t]
\centering
\begin{tikzpicture}[
  role/.style={draw, rounded corners, minimum width=2.0cm, minimum height=0.7cm, align=center,
    font=\footnotesize},
  state/.style={draw, thick, rounded corners, minimum width=2.3cm, minimum height=1.1cm,
    align=center, font=\footnotesize\bfseries, fill=gray!12},
  arr/.style={-{Latex[length=1.8mm]}, thick}
]
\node[state] (state) at (0,0) {Navigation\\State};
\node[role] (sens) at (-2.9, 0.0) {Sensing};
\node[role] (map)  at (2.9, 0.0) {Mapping /\\Representation};
\node[role] (loc)  at (-2.9, -1.4) {Localization};
\node[role] (plan) at (2.9, -1.4) {Planning};
\node[role] (ctrl) at (0, -1.4) {Control};
\draw[arr] (sens) -- (state);
\draw[arr] (map) -- (state);
\draw[arr] (state) -- (loc);
\draw[arr] (state) -- (plan);
\draw[arr] (state) -- (ctrl);
\draw[arr] (loc) -- (state);
\draw[arr] (plan) -- (state);
\draw[arr] (ctrl) -- (state);
\end{tikzpicture}
\caption{Functional roles interact exclusively through the shared, time-stamped navigation state;
no direct dependencies exist between roles.}
\label{fig:arch:overview}
\end{figure}

The proposed architecture organizes navigation around a strict separation of concerns and explicit
control of data flow, execution, and time: functional roles interact only through well-defined
interfaces and shared state (Figure~\ref{fig:arch:overview}), enabling modularity, reusability, and
principled reasoning about timing and consistency.

\subsection{System Organization and Execution Model}

The pipeline decomposes into a small set of functional subsystems, sensing, environment
representation, state estimation, planning, and control, each interacting with the rest exclusively
through the shared navigation state, with no direct dependencies between them and no assumptions on
intermediate representations (e.g., planners need not output purely geometric paths). Not every
role is required in every deployment; trivial or placeholder components may stand in when unneeded,
letting the same structure span from purely reactive motion to global navigation.

\subsection{Modular Components and Representation Independence}

A central goal is to decouple navigation functionality from environment representation: environment
models are interchangeable components that may coexist, cooperating sequentially or serving
different consumers (e.g., planar maps, multi-level surfaces, volumetric models), with
representation-specific processing, such as obstacle integration or safety margins, encapsulated
within them so it never leaks into localization, planning, or control, which instead operate over
abstract navigation data and remain interchangeable across sensing setups as long as they adhere to
the navigation state interface.

\subsection{Navigation State as an Explicit Coordination Mechanism}

At the core lies an explicit navigation state $\mathcal{S}$, accessed exclusively by navigation
components and organized as a set of uniquely keyed entries $e=(k,v,t,f)$: a string key $k$, a value
$v$ of arbitrary type, its generation timestamp $t$, and the coordinate frame $f$ it refers to when
applicable. Let $K$ denote the set of all keys that may appear in $\mathcal{S}$ across the
navigation pipeline, i.e., the union of the keys declared by every component. Each component
independently declares the keys it requires, $R \subseteq K$, and produces, $W \subseteq K$, and may
only read $R$ and write $W$. A pipeline is well formed only if
every required key is produced by some active component, turning compatibility into an explicit,
inspectable requirement rather than an implicit assumption. The reference implementation documents
this contract through normalized per-plugin specification sheets (Section~V); automatically
checking it at assembly time is left as future tooling.

Because entries are uniquely keyed rather than overwritten, $\mathcal{S}$ lets components cooperate
or compete over multiple coexisting estimates without ambiguity, e.g., different representations for
prediction versus correction, or planning versus collision checking, enabling systematic reasoning
about data validity, consistency, and timing.

\subsection{Explicitly Ordered Execution and Reactive Scheduling}

Components are associated with nominal execution rates that bound computational activity, skipping
computation when insufficient time has elapsed without affecting pipeline consistency. Upstream
components can also explicitly signal new information, temporarily overriding nominal rates
downstream to improve end-to-end reaction time; responsiveness thus becomes an explicit scheduling
property rather than a side effect of asynchronous execution, avoiding reliance on implicit callback
timing or middleware-specific scheduling that obscures latency.

\subsection{Time and Frame Consistency as a Design Principle}
\label{sec:arq:tf}

The architecture treats temporal consistency as a fundamental concern rather than an implementation
detail: navigation decisions are only meaningful when the validity times of the data and
transformations they combine are explicit and satisfy the temporal-consistency policy required by
the consumer; ignoring this leads to systematic geometric errors and compromises determinism,
particularly in dynamic environments or systems with non trivial kinematics. Concretely, this
guarantee holds per entry, via the timestamp $t$ attached to every value and explicit freshness
checking by consumers, not via atomic cross-key snapshots or an assumption of simultaneity: two
entries read together, e.g., pose and path, may originate from different write cycles, each still
carrying the timestamp needed to reason about its own validity.

All navigation data carries a validity timestamp, inherited from the physical instant it refers to,
distinct from the wall-clock instant a component computes and emits its output. Estimates inherit
the validity timestamp of their sensory or odometric sources even if computed later, and
trajectories and control commands are tagged with the validity timestamp of their most recent
contributing input, so freshness reasoning reflects when data is valid for, not when it was
processed; their separate emission time, when a command is actually published, is what end-to-end
latency measurements use (Section~V).

Coordinate transformations obey the same constraint: they are queried using the timestamp of the
data they apply to, and substituting the current system time for correct temporal labeling is an
architectural error except when explicitly justified, since it introduces delay and geometrically
inconsistent results. Transformations are therefore managed through a dedicated, centrally
accessible store guaranteeing immediate availability once produced, avoiding asynchronous delivery
that could introduce hidden latency.

Internally generated transformations are inserted directly into this store, so subsequent components
access them deterministically within the same cycle, while external transformations go through a
time-aware, non-blocking ingestion mechanism that preserves ordering. Transformation availability
thus becomes an explicit architectural property rather than an emergent side effect of middleware
scheduling, a foundation for predictable, robust navigation behavior often absent from existing
stacks.

\section{Reference Implementation}
\label{sec:impl}

This section demonstrates that the architecture of Section~III can be realized within a realistic
robotic system, \emph{EasyNav}, under practical constraints such as real-time execution,
transformation handling, and integration with existing robotic software ecosystems, without
introducing additional architectural concepts. Although the architecture itself is middleware
agnostic, the reference implementation is realized in ROS~2 to demonstrate feasibility under
realistic conditions and facilitate reproducibility.

\subsection{Components and plugins}
\label{sec:impl:components}
\begin{figure*}[t]
\centering
\includegraphics[width=0.78\linewidth]{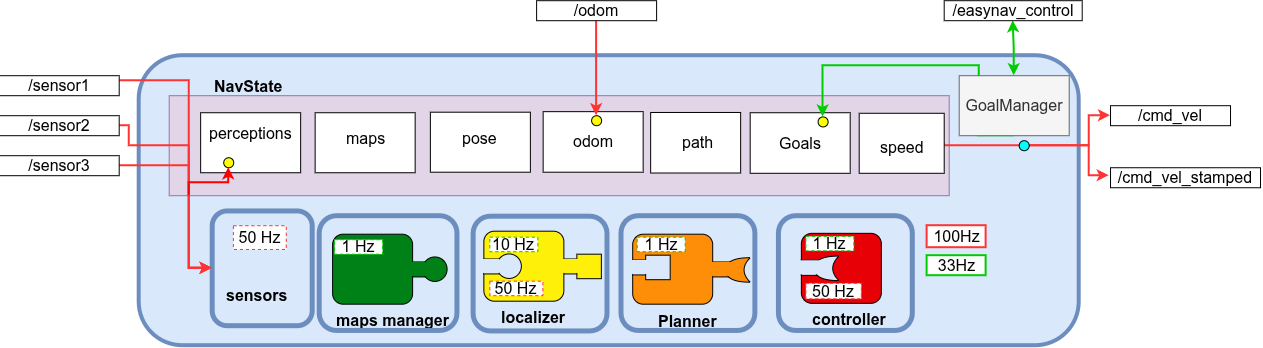}
\caption{EasyNav Design. The \textbf{NavState} blackboard (lavender box) stores the information
that plugins read and write. \textbf{Sensors} is the single entry point through which sensory data
enters the system. Colored puzzle pieces represent the plugin placeholders of each functional role,
which must fit together to form a valid pipeline. Navigation goals reach the system through the
\textbf{GoalManager}, which writes them into NavState for the Planner to consume. Red boxes indicate
execution frequencies in the real-time cycle, and green boxes indicate execution frequencies in the
non-real-time cycle.}
\label{fig:easynav:design}
\end{figure*}

EasyNav is a single monolithic process built around a ROS~2 Lifecycle Node, the \emph{System Node},
which internally contains five ROS~2 Lifecycle Nodes (Figure~\ref{fig:easynav:design}) that load and
manage the plugins implementing each navigation stage and activate their processing cycles at
configured rates. The specific plugins, their execution rates, and their parameters are all
specified through a single configuration file, enabling reproducible deployment and explicit control
over system behavior without code changes. The functional role of each node is summarized below:

\begin{itemize}
  \item \textbf{Sensors Node.} This node is the single entry point through which sensory
  information enters the navigation state; plugins elsewhere may still create auxiliary
  subscriptions for monitoring, but these do not participate in inter-component navigation data
  flow. Each sensor
  is handled by a dedicated perception-handler plugin, uniformly covering point-based sensors
  (\texttt{LaserScan}, \texttt{PointCloud2}), images, inertial data, GPS, odometry, and external
  detections; new sensor types are supported by registering an additional handler, without
  modifying the Sensors Node.

  \item \textbf{Maps Manager Node.} This node manages environment representations, or maps, through
  one or more map-manager plugins, configured and updated sequentially so that one plugin may
  consume another's output; each can be further structured into sub-plugins, e.g., obstacle
  filtering or safety-margin inflation. This supports, for example, outdoor setups where a
  navigable-surfaces representation feeds localization's prediction stage and the planner, while a
  volumetric representation feeds the correction stage, and a route plugin that biases a
  costmap-based planner toward predefined waypoint graphs while still avoiding obstacles
  (Figure~\ref{fig:intro:fig3}). As a further illustration of representation independence, EasyNav
  has also been deployed with \emph{NavMap}, a map-manager plugin that represents navigable surfaces
  as a triangular mesh supporting multi-surface environments (Figure~\ref{fig:intro:fig1}),
  requiring no modification to the remaining components.

    \item \textbf{Localizer Node.}
    This node instantiates a localization plugin that estimates robot position from odometry,
    perceptual data, and one or more environment representations, explicitly supporting less
    conventional outputs too, e.g., pose sets with probabilities, probabilistic grids, or
    topological regions; the only requirement is that the chosen planner or controller can consume
    whichever output the localizer produces.

    \item \textbf{Planner Node.} This node hosts a planning plugin that computes a route to a goal.
    While the common case is a sequence of poses, the architecture does not constrain the output
    format: alternative representations, e.g., waypoints annotated with timestamps or velocities,
    enable advanced behaviors such as social-navigation speed adaptation to avoid interfering with
    nearby people.

    \item \textbf{Controller Node.}
    This node instantiates a control plugin that generates velocity commands, using both stamped
    and non-stamped twist representations, the standard motion-command interfaces under ROS~2.

\end{itemize}
 
To implement the navigation state defined by the proposed architecture, a thread-safe blackboard
mechanism has been developed. This blackboard allows arbitrary data types to be stored and retrieved
using string-based keys, enabling flexible information exchange between navigation components.
Throughout the remainder of this paper, including the algorithms below, this blackboard object is
referred to by its implementation name, \textbf{NavState}.

Special care avoids continuous dynamic memory allocation, incompatible with real-time execution:
allocation occurs only when a new entry is created, and subsequent accesses update data via copy
operations. Perceptual data, updated at high frequency, is the most critical case; since a given
sensor's data typically has constant size throughout execution, data structures optimized for this
pattern prevent performance degradation.

The blackboard is owned by the System Node and a reference to it is passed to each internal node
whenever its processing cycle is executed. This design ensures a single shared navigation state
while avoiding unnecessary data duplication and preserving deterministic access patterns.

Individual reads and writes are serialized through a single internal mutex, giving atomic per-entry
access without atomic cross-key snapshots, consistent with Section~\ref{sec:arq:tf}. For entries
written from a different thread than the one reading them, e.g., path, goals, or pose across the
real-time/non-real-time boundary, a safe-copy accessor avoids exposing a concurrently overwritten
value to the reader; components that both write and read the same key from one thread use a plain
reference accessor instead, avoiding an unnecessary copy.

\subsection{Managing Time and Geometry}
\label{sec:impl:tf}

In ROS~2, the transformation system maintains geometric relationships between reference frames over
time, following REP~105\footnote{\url{https://www.ros.org/reps/rep-0105.html}}, strictly enforced
in our implementation: updates (TFs) are published on \texttt{/tf} and \texttt{/tf\_static} and
received through a buffer fed by a listener, which resolves queries between arbitrary frames by
interpolating between the closest available samples, waiting up to a configurable timeout, or
erroring when the request would require extrapolating into the future. This standard mechanism is
incompatible with our execution model, where transformations may be produced and consumed within
the same system cycle: waiting for a transformation to be published and then processed by a
listener callback is not acceptable.

For this reason, our architecture employs a dedicated real-time transformation buffer, the
\textbf{RTTFBuffer}: it is fed by a transformation listener running in its own real-time thread,
dedicated exclusively to external transformations; all internally produced transformations are
inserted directly into it at creation time, in addition to being published through a broadcaster;
and it is globally accessible to all nodes and plugins through a singleton interface. The system
further adheres to the temporal labeling criteria of Section~\ref{sec:arq:tf} and never waits for a
transformation to become available at a given timestamp, since doing so would introduce uncertainty
and compromise execution predictability.

There are two situations in which a required transformation may not be available at the exact time
it is queried:

\begin{itemize}
  \item A sensory measurement from sensor $s$, whose associated coordinate frame is denoted $F_s$,
  typically point-based data such as 2D or 3D range scans, is acquired at time $t_i$, but the
  transformation from the robot base frame to $F_s$ is not available until time $t_j$, while the
  transformation is required at time $t_k$, with $t_i < t_k < t_j$. In this case, a perception
  buffer is used, and the most recent valid measurement whose timestamp is less than $t_k$ is
  selected. Given that the transformation listener operates in real time, this situation only
  occurs under exceptional circumstances.
  \item The maps manager needs, for a perception acquired at time $t_p$, the transformation from
  $F_s$ to the map frame. This transformation is produced by localization from its own,
  potentially earlier, perception at time $t_\ell \leq t_p$, since localization executes far more
  frequently, in the real-time thread, than the maps manager, which runs in the non-real-time
  thread; the maps manager accepts the transformation computed at $t_\ell$ instead of waiting for
  one matching $t_p$ exactly. This is an explicit bounded-staleness policy, not an unquantified
  approximation: the temporal error $t_p - t_\ell$ is bounded by localization's execution period in
  the real-time thread (e.g., 20~ms at its nominal 50~Hz rate), since a fresher transformation
  becomes available every cycle; the maps manager, updated at a lower rate, simply consumes the
  most recent one.
\end{itemize}

\subsection{Real-Time Execution Model}

As discussed, the framework executes as a single monolithic process launched from a single
configuration file, avoiding complex interdependencies between multiple launcher files and
simplifying deployment, debugging, and reproducibility.

Within this process, two threads execute the navigation cycles, one real-time and one non-real-time;
the RTTFBuffer's transformation listener runs in the separate dedicated real-time thread described
in Section~\ref{sec:impl:tf}. The navigation-cycle real-time thread is configured to use FIFO
scheduling with a high priority, typically
around 80, and runs under the PREEMPT\_RT fully preemptible kernel configuration, natively available
in mainline Linux without requiring an out-of-tree patch. Neither thread delegates its execution
cycle or frequency control to the ROS~2 executor's timer dispatch: each thread governs its own
wall-clock rate loop directly, invoking the corresponding system cycle every iteration and spinning
its callback group only for a bounded duration. This avoids the default ROS~2 executor, which
imposes no explicit callback execution order and provides no built-in triggering control, and can
result in priority inversion and starvation of ready callbacks under high load or tightly
constrained scheduling conditions\footnote{
\url{https://docs.ros.org/en/rolling/Concepts/Intermediate/About-Executors.html}}. In such
situations, the resulting behavior can negatively affect the intended execution frequencies. At
startup, the shared \textbf{NavState} and \textbf{RTTFBuffer} are created before the real-time and
non-real-time threads are spawned, each running its own loop that invokes the system's real-time or
non-real-time cycle, detailed in Algorithm~\ref{alg:easynav_cycles}, while the system is active; on
shutdown, both threads are signaled to stop and joined before the navigation system itself is shut
down.

Instead of timer-driven execution, the system explicitly manages execution cycles and degrades
gracefully when nominal rates cannot be sustained, favoring predictable behavior over implicit
scheduling; limiting the thread count further improves resource-usage determinism, an important
property when navigation coexists with other time-critical processes.

The System Node propagates execution to each component's \texttt{update()} or \texttt{update\_rt()}
method, both returning immediately once their configured frequency has not yet been reached. The
real-time cycle additionally lets upstream components trigger execution ahead of their nominal
rate: new sensor data (line 2) or a newly updated robot pose from localization (line 3) each set a
trigger flag that propagates downstream, temporarily overriding nominal frequencies to improve
reaction time to dynamic events, e.g., an obstacle's sudden appearance, while preserving a single,
explicitly ordered flow, as Algorithm~\ref{alg:easynav_cycles} illustrates.

\begin{algorithm}[h!]
\footnotesize
\caption{System execution cycles, trigger propagation, and frequency gating in EasyNav}
\label{alg:easynav_cycles}
\begin{algorithmic}[1]

\STATE \textbf{Real-Time Cycle} \textsc{system\_cycle\_rt}(\textbf{NavState})
\STATE $\textit{trigger\_perceptions} \leftarrow$ Sensors.cycle\_rt(\textbf{NavState})
\STATE $\textit{trigger\_localization} \leftarrow$ Localizer.cycle\_rt(\textbf{NavState},
$\textit{trigger\_perceptions}$)
\STATE $\textit{trigger} \leftarrow \textit{trigger\_perceptions} \lor
\textit{trigger\_localization}$
\STATE $\textit{trigger\_controller} \leftarrow$ Controller.cycle\_rt(\textbf{NavState},
$\textit{trigger}$)
\COMMENT{Execution may be skipped if frequency gating is not satisfied and no trigger is active}

\IF{\textbf{NavState} contains \texttt{cmd\_vel} \AND $\textit{trigger\_controller}$}
  \STATE Publish motion command from \textbf{NavState}
  \COMMENT{stamped or non-stamped, depending on configuration}
\ENDIF

\vspace{0.6em}

\STATE \textbf{Non-Real-Time Cycle} \textsc{system\_cycle}(\textbf{NavState})
\STATE Sensors.cycle(\textbf{NavState})
\STATE Localizer.cycle(\textbf{NavState})
\STATE MapsManager.cycle(\textbf{NavState})
\STATE GoalManager.update(\textbf{NavState})

\STATE $\textit{goal\_updated} \leftarrow$ (GoalManager.timestamp $>$
Planner.last\_execution\_timestamp)
\STATE Planner.cycle(\textbf{NavState}, $\textit{goal\_updated}$)
\COMMENT{Planner execution is subject to frequency gating unless goals have changed}

\IF{NavState publication is enabled}
  \STATE Publish debug view of \textbf{NavState}
\ENDIF

\end{algorithmic}
\normalsize
\end{algorithm}

\section{Experimental Evaluation}
\label{sec:eval}
The objective of the experimental evaluation is to validate the architectural principles proposed in Section~\ref{sec:arch} under realistic robotic conditions, real robots, and real workloads, focusing on two consequences directly tied to the execution and data-flow mechanisms introduced by the architecture: computational resource consumption and navigation reaction latency. Nav2 is used throughout as the baseline against which these consequences are measured, not as the subject of the evaluation: it is the most widely adopted ROS~2 navigation stack, so comparing against it under matched conditions is what makes the practical effect of the architectural choices in Section~\ref{sec:arch}, an explicit shared state, timing and frame consistency, and perception-triggered execution, observable and measurable.

\subsection{Experimental Setup}
\label{sec:eval:setup}

\paragraph{Platforms and pipelines}

Experiments are conducted on two mobile robotic platforms representing complementary navigation scenarios, summarized in Table~\ref{tab:platforms}. Both platforms are evaluated with EasyNav~0.4.0 and Nav2~1.3.12 under the corresponding symmetric configuration; Nav2 is deployed as a composed set of nodes with intra-process communication~\cite{macenski2023impactros2node} enabled, providing an execution model close to the component-based EasyNav architecture.

The indoor scenario uses a TurtleBot~4 navigating a structured environment with narrow passages and static obstacles using a 2D planar costmap; both frameworks run on the same onboard Raspberry~Pi~4 hardware, configured in performance mode, under ROS~2 Jazzy with \texttt{Fast DDS}. The outdoor platform, a Summit~XL (Figure~\ref{fig:intro:fig4}), traverses a university campus route using GNSS and IMU sensing over the same costmap-based representation. In both cases, navigation components are composed through the respective framework interfaces without modifying their implementations.

\paragraph{Symmetric configuration procedure}
To ensure a fair comparison, configurations are adapted symmetrically with respect to the reference system, reversing the adaptation direction between indoor and outdoor experiments (Table~\ref{tab:platforms}). Indoors, EasyNav is adapted to reproduce, as closely as possible, Nav2's navigation behavior and operating conditions. However, deploying Nav2 on the resource-constrained Raspberry~Pi~4 required removing non-essential nodes outside the evaluated pipeline and subscriptions to topics from other machines. The Nav2 lifecycle heartbeat period was also increased from the reference 0.1~s to 0.5~s, following measurements reported in the Nav2 issue discussion\footnote{\url{https://github.com/ros-navigation/navigation2/issues/5784##issuecomment-3663491309}}, which showed substantially lower CPU utilization with longer heartbeat intervals; this parameter only affects lifecycle supervision.

Outdoors, conversely, the Nav2 configuration is adapted to reproduce the navigation behavior and operating conditions of EasyNav. In both indoor and outdoor scenarios, the frameworks employ identical plugins or equivalent algorithmic ports. Parameters are matched according to their functional role whenever a direct correspondence exists, while implementation-specific parameters are retained otherwise; full parameter mappings are provided in the paper repository\footnote{\ifanon Repository omitted for anonymous review.\else
\url{https://github.com/EasyNavigation}\fi}. All adaptations are defined before data collection and kept fixed across runs; neither system is subsequently tuned based on the measured results.

\paragraph{Latency experiment configuration}
The latency experiment reuses the configuration above with two changes. Indoors, EasyNav's localization update frequency is set to 2~Hz (periodic, via its \texttt{freq} parameter), matching the effective update frequency measured for Nav2's event-driven AMCL, which has no direct frequency equivalent; and Nav2's lifecycle heartbeat is increased further, from 0.5~s to 0.65~s, following the same rationale as the main configuration (Section~\ref{sec:eval:setup}) to offset the extra lifecycle-supervision overhead from the blocking experiment (decided before data collection and conservative toward Nav2, not EasyNav). Outdoors, Nav2's real-time components (e.g., the controller) are set to $60$~Hz and its non-real-time components (e.g., costmaps) to $5$~Hz, mirroring EasyNav's two-cycle design.

\begin{table}[htbp]
\centering
\scriptsize
\caption{Experimental platforms and navigation pipelines.}
\label{tab:platforms}
\begin{tabular}{p{1.9cm}p{1.9cm}p{2.7cm}}
\hline
& \textbf{Indoor} & \textbf{Outdoor} \\
\hline
Platform & TurtleBot~4 & Summit~XL \\
Hardware & Raspberry Pi~4 & i7-10700 \\
Middleware & Fast DDS & Fast DDS + Zenoh bridge \\
Sensing & LiDAR 2D & LiDAR 3D + IMU + GNSS \\
Env.\ repr. & Costmap & Costmap \\
Localization input & Range + odom. & IMU + odom. + GNSS \\
\hline
\end{tabular}
\end{table}

\subsection{Metrics and Measurement Methodology}
\label{sec:eval:metrics}

\paragraph{Computational resource consumption (indoor and outdoor)}

Indoors, the navigation process is sampled every 50~ms per execution. CPU utilization is the change in process user and system CPU time between samples, normalized by elapsed time and core count; memory is the resident set size reported by the OS. Command and controller update frequencies come from message timestamps, and travelled distance from accumulated odometry displacement. Each complete execution is the experimental unit: measurements are aggregated within each run before the resulting run-level observations are compared, avoiding treating temporally correlated samples as independent.

Outdoors, EasyNav and Nav2 run inside a Docker container isolated on four dedicated cores locked at maximum frequency; CPU and memory are obtained from Linux \texttt{cgroups} (\texttt{anon} memory, to avoid disk-cache skew). Unlike indoors, these data come from a single, extensive continuous trial per framework (Section~\ref{sec:eval:outdoor}) rather than multiple independent executions; the reported mean and standard deviation therefore describe sample-level variability within that one deployment, not between-run variability, and are treated as descriptive of long-duration behavior rather than an inferential comparison.

Beyond resource consumption, the study also reports \texttt{cmd\_vel} publication frequency, controller
%mean minimum LiDAR obstacle distance,
update frequency, and travelled distance. The gap between publication and update frequency is informative: it reveals whether a fresh command is computed every cycle or a previous one republished to sustain the nominal rate, directly reflecting the freshness principle of Section~\ref{sec:arq:tf}. Travelled distance instead characterizes how each framework executes the same task, not a standalone measure of navigation quality or safety, which depends on the specific algorithms and configurations used.

\paragraph{Navigation reaction latency (indoor and outdoor)}
Reaction latency is defined as $T_{\mathrm{latency}} = T_{\mathrm{stop}} - T_{\mathrm{trigger}}$, where
$T_{\mathrm{trigger}}$ is the timestamp at which a controlled perceptual-blocking event is received by
the navigation stack, and $T_{\mathrm{stop}}$ is the timestamp of the first \texttt{cmd\_vel} message
satisfying the configured stopping thresholds.

\subsection{Results}
\subsubsection{Computational Efficiency}
\label{sec:eval:rq1}
\paragraph{Indoor Navigation}
\label{sec:eval:indoor}

The robot performs a patrol mission of three laps through five predefined waypoints. Each execution
lasts approximately nine minutes and covers approximately 130~m; ten independent executions are
performed per framework.

\begin{figure}[hbtp]
  \centering
  \includegraphics[width=\linewidth]{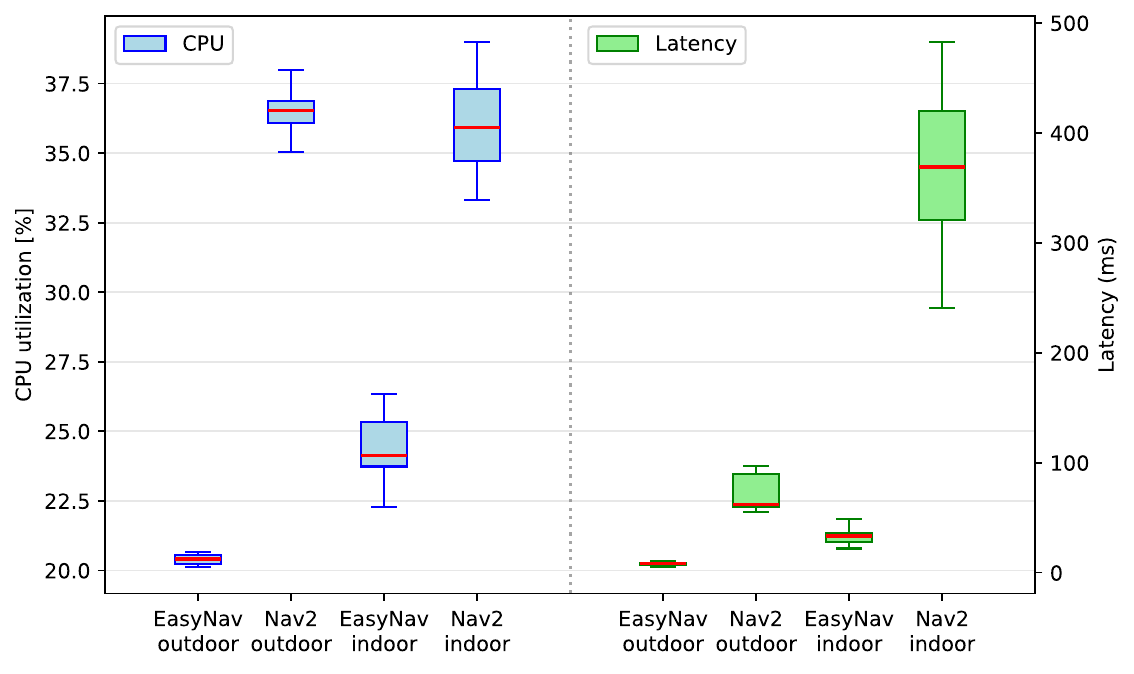}
  \caption{Comparison of CPU utilization and latency on both experiments. Outdoor CPU boxes
  summarize within-trial samples from a single continuous run, not independent executions
  (Section~\ref{sec:eval:metrics}).}
  \label{fig:cpu_lat_results}
\end{figure}

EasyNav exhibits substantially lower computational resource consumption than Nav2
(Table~\ref{tab:results}, Figure~\ref{fig:cpu_lat_results}). Despite the reduction of non-essential
components required to run Nav2 on the Raspberry~Pi~4, Nav2 exhibits substantially higher CPU
utilization than EasyNav. Mean CPU utilization is 24.38\% for
EasyNav versus 36.01\% for Nav2, corresponding to a mean reduction of $11.63$ percentage points.
Mean memory consumption is 115.83~MB for EasyNav and 189.96~MB for Nav2, a reduction of
$74.13$~MB.

To characterize execution behavior beyond resource consumption, we compare command publication, controller update, and trajectory statistics. EasyNav publishes \texttt{cmd\_vel} at $49.41$~Hz, substantially higher than Nav2's $18.11$~Hz, closely matching its $50$~Hz real-time cycle; the controller itself has a nominal update frequency of $20$~Hz but is observed at $23.02$~Hz, since EasyNav evaluates the controller every real-time cycle while frequency gating governs when a new computation actually occurs, letting the most recent command be republished without a new controller update. Nav2's observed controller update frequency is lower, at $16.23$~Hz. EasyNav also accumulates a longer travelled distance ($142.23$~m vs. $132.09$~m), reflecting the distinct closed-loop trajectories produced by each framework under matched patrol conditions rather than a standalone measure of navigation quality; no collisions were observed in any indoor execution.

\paragraph{Outdoor Navigation}
\label{sec:eval:outdoor}

Unlike the indoor evaluation, the outdoor assessment consists of a single continuous $2.75km$ trial rather than multiple discrete runs, during which the platform repeatedly navigates several laps along a predefined route across the university campus. This continuous methodology assesses long-term stability, memory management, and computational endurance under sustained deployment conditions.

Both frameworks successfully completed the entire circuit. The collected data demonstrate a substantially lower resource footprint for EasyNav, showing an average CPU utilization of $20.23\%$ and $93.83$ MB of RAM, compared to the $37.64\%$ CPU and $249.7$ MB of RAM consumed by Nav2.

Regarding command generation, both frameworks satisfy the nominal $60$~Hz publication rate. Crucially, only EasyNav sustains this frequency in actual controller updates ($59.92$Hz), computing a new velocity command from the latest available navigation state at every cycle. In contrast, Nav2 drops to an update frequency of $33.14 $~Hz, regularly republishing repeated commands to sustain the nominal publication rate.

\subsubsection{Navigation Reaction Latency}
\label{sec:eval:rq2}
We evaluate reaction latency with controlled indoor and outdoor experiments. A \texttt{scan\_mode\_bridge} node is interposed
between the LiDAR and the navigation stack, forwarding measurements unchanged during normal operation
and, at a predefined point, switching to a blocked mode that replaces LiDAR ranges with a fixed
blocked value, providing an identical controlled trigger to both frameworks. Twenty independent
executions are performed per framework and platform; in each, the robot navigates toward a predefined goal for a
fixed period before the blocking event is triggered.

EasyNav reacts markedly faster than Nav2 in this setting, with a mean latency of 32.97~ms versus
377.19~ms for indoor experiment and 7.86~ms versus
70.68~ms for outdoor experiment  (Table~\ref{tab:results}).

\begin{comment}
\begin{figure}[t]
  \centering
  \begin{subfigure}[t]{0.75\linewidth}
    \centering
    \IfFileExists{fig/latency_distribution_indoor.png}{%
      \includegraphics[width=\linewidth]{fig/latency_distribution_indoor.png}%
    }{%
      \fbox{\parbox[c][3.4cm][c]{0.9\linewidth}{\centering latency\_distribution.png}}%
    }
    \caption{Run-level reaction latency after the controlled blocking trigger.}
    \label{fig:rq2_latency_dist}
  \end{subfigure}
  \caption{Reaction latency in the controlled perceptual-blocking experiment. EasyNav stops
    substantially sooner than Nav2 after the same trigger event.}
  \label{fig:rq2_latency}
\end{figure}
\end{comment}

\subsection{Discussion}
\label{sec:eval:discussion}
Table~\ref{tab:results} summarizes the quantitative findings, organized by scenario.

\begin{table}[hbtp]
  \centering
  \scriptsize
  \renewcommand{\arraystretch}{0.85}
  \setlength{\tabcolsep}{3pt}
   \caption{Summary of the experimental results. Values are reported as mean $\pm$ standard
    deviation across independent runs, except for outdoor CPU/memory/frequency rows, where they
    describe sample-level variability within a single continuous trial
    (Section~\ref{sec:eval:metrics}).}
  \label{tab:results}
  \begin{tabular}{p{0.8cm}p{2cm}r r r}
    \hline
    \textbf{Scenario} & \textbf{Metric} &
    \textbf{EasyNav} & \textbf{Nav2} & \textbf{Diff.} \\
    \hline

    \multirow{6}{*}{Indoor}
    & CPU [\%] &
    $24.38 \pm 1.35$ & $36.01 \pm 1.81$ & $-11.63$~pp \\

    & Memory [MB] &
    $115.83 \pm 0.60$ & $189.96 \pm 0.65$ & $-74.13$ \\

    & Cmd freq. [Hz] &
    $49.41 \pm 0.71$ & $18.11 \pm 0.87$ & $+31.30$ \\

    & Cmd ud. freq. [Hz] &
    $23.02 \pm 3.03$ & $16.23 \pm 1.92$ & $+6.79$ \\

    & Travelled dist. [m] &
    $142.23 \pm 0.67$ & $132.09 \pm 0.44$ & $+10.14$ \\

    & Latency [ms] &
    $32.97 \pm 7.20$ & $377.19 \pm 80.16$ & $-344.21$ \\

    \hline

    \multirow{6}{*}{Outdoor}
    & CPU [\%] &
    $20.23 \pm 1.03$ & $37.64 \pm 5.51$ & $-17.41$~pp \\

    & Memory [MB] &
    $93.83 \pm 0.04$ & $249.70 \pm 0.39$ & $-156$ \\

    & Cmd freq. [Hz] &
    $59.92 \pm 0.34$ & $59.97 \pm 0.25$ & $-0.05$ \\

    & Cmd ud. freq. [Hz] &
    $59.92 \pm 0.34$ & $33.14 \pm 20.58$ & $+26.78$ \\

    & Travelled dist. [km] &
    $2.75$ & $2.75$ & $0$ \\

    & Latency [ms] &
    $7.86 \pm 1.94$ & $70.68 \pm 16.66$ & $-62.82$ \\

    \hline
  \end{tabular}
\end{table}

Regarding computational efficiency, EasyNav reaches the same indoor patrol objective with lower CPU
and memory demand than Nav2, exhibiting denser velocity-command publication and longer
trajectories. Notably, this
indoor comparison already favors Nav2: reaching a viable configuration on the resource-constrained
Raspberry~Pi~4 required trimming Nav2 down to the nodes used by the evaluated pipeline
(Section~\ref{sec:eval:setup}), reducing its footprint relative to a default deployment, yet
EasyNav still exhibits substantially lower CPU and memory usage. The outdoor evaluation extends
the comparison to a different robot, sensing stack, and adaptation direction, confirming the
framework's reliability across platforms. Both platforms use the same costmap-based representation
(Table~\ref{tab:platforms}); heterogeneous representations coexisting in a single pipeline are
instead demonstrated qualitatively via the NavMap integration (Section~\ref{sec:impl:components}),
with quantitative characterization left to future work.

Regarding reaction latency, both experiments show that EasyNav achieves substantially lower reaction latency than Nav2 in the controlled perceptual-blocking experiment, consistent with the perception-triggered execution mechanism, in which newly received sensory data explicitly activates the ordered real-time execution path instead of waiting for the next controller-driven update. In the same indoor configuration, the measured real-time cycle frequency (49.99~Hz) closely matches both its nominal rate and the observed \texttt{cmd\_vel} publication frequency (49.41~Hz), consistent with commands being published at the cycle frequency while controller recomputation remains independently gated.

\section{Conclusion}
\label{sec:conclusion}

This paper presented a navigation architecture organized around an explicit, time-stamped
navigation state as the sole interaction interface between mapping, localization, planning, and
control, validated through EasyNav, a fully open source ROS~2 realization deployed on real indoor
and indoor-outdoor platforms. Under a matched comparison against Nav2, EasyNav consistently uses
less CPU and memory while sustaining higher controller-update frequencies and maintaining the
configured command-publication rates, and reacts to a controlled perceptual-blocking event with
both a lower mean and a lower spread in reaction latency, consistent with a perception-triggered,
explicitly ordered execution path rather than implicit, timer-driven scheduling. These results
support the claim that treating navigation state, timing, and frame consistency as explicit
architectural properties is compatible with practical real-time performance, without trading away
the modularity and representation independence central to the proposed architecture. Two
limitations follow directly from the scope of this evaluation: the individual contribution of the
perception-triggered execution mechanism is not yet isolated through targeted ablations, and the
coexistence of heterogeneous representations, demonstrated qualitatively through the NavMap
integration, is not yet characterized quantitatively; both are left as future work.

% Hidden for double-anonymous review: grant numbers can deanonymize authors/institution.
\ifanon\else
\section*{ACKNOWLEDGMENT}
This work was supported by the AEI under grant PID2024-161761OB-C22 (AURORAS), the EU Horizon Europe program under grant 101070254 (CORESENSE), and the Community of Madrid under grant TEC-2024/TEC-62 (iRoboCity2030-CM).
\fi

\bibliographystyle{IEEEtran}
\bibliography{references}

\end{document}